\documentclass[letterpaper,10pt,conference]{ieeeconf}
\IEEEoverridecommandlockouts
\usepackage{times}
\usepackage{mathptmx}
\usepackage{amsmath,amssymb}
\usepackage{array}
\usepackage{booktabs}
\usepackage{graphicx}
\usepackage{cite}
\usepackage{etoolbox}
\usepackage[hidelinks]{hyperref}

\makeatletter
\patchcmd{\@makecaption}{\scshape}{\upshape}
  {}{\PackageError{arxiv-layout}{Table caption style patch failed}{Check the caption definition in ieeeconf.cls.}}
\renewcommand{\fnum@table}{Table~\thetable}
\makeatother

\title{\LARGE\bfseries A Lightweight Hybrid Approach for Milling Force Reconstruction from Internal Machine Signals}

\author{Eric Hirsch$^{1,*}$, Christian Friedrich$^1$%
\thanks{$^1$Institute for Robotics and Intelligent Production Systems, Karlsruhe University of Applied Sciences (HKA), 76133 Karlsruhe, Germany. $^*$Corresponding author. E-mail address: eric.hirsch@h-ka.de.}%
}
\specialpapernotice{Preprint}

\begin{document}
\maketitle
\thispagestyle{plain}
\pagestyle{plain}
\raggedbottom

\begin{abstract}
Data-driven monitoring based on milling force can enable advanced manufacturing services including process optimization and condition monitoring. However, industrial implementation is often hindered by costly and intrusive hardware and large training data requirements. This paper therefore presents a hybrid approach using machine-internal signals and machining metadata to enable force-based monitoring on brownfield machines without additional sensors. Commonly available machine signals have relatively low sampling rates and cannot directly represent high-frequency milling-force variations, especially at higher spindle speeds. Instead of learning the complete force signal, which would increase model complexity, data requirements, and overfitting risk, a supervised machine-learning model predicts only the average milling force from selected machine-signal features and machining parameters. This low-frequency estimation parameterizes a static milling force model that reconstructs the force behavior over one spindle revolution. Although the static model does not reproduce all dynamic effects, it captures the force distribution, load levels, and angular regions of highest tool engagement. The approach is evaluated on one milling machine across varying process parameters, engagement conditions, cutting tools, and two workpiece materials. Dynamometer measurements are required only for development and validation. Ridge regression achieves an $R^2$ of $0.984$ and $0.991$ for average-force prediction on the test data ($\bar{F}_x$ and $\bar{F}_y$). The reconstructed force behavior over one spindle revolution achieves an $R^2$ of $0.962$ and $0.982$ ($F_x$ and $F_y$) compared with the reconstruction based on measured force features. The method provides a cost and computationally efficient basis for process monitoring, tool condition assessment, and process optimization on existing machines.
\end{abstract}


\section{Introduction}
\label{sec:introduction}

\noindent Cutting forces reflect the tool–workpiece interaction and provide valuable information for process monitoring and optimization~\cite{ochudlo_model_2026,liu_monitoring_2022}.
Their magnitude and progression depend on process conditions, machining system dynamics, and tool state. Process conditions determine the nominal chip thickness, while relative motion between the tool and workpiece modifies it. The tool state influences how this chip thickness translates into force. These forces, in turn, cause deflections and vibrations that further change the chip thickness. This feedback makes force prediction across operating conditions challenging.

Consequently, laboratory studies commonly use piezoelectric dynamometers for direct force measurement~\cite{ni_mechanism_2026,ebrahimi_araghizad_improving_2024, dib_review_2026}. However, their cost, installation effort, and workspace constraints can hinder routine industrial use~\cite{yedurkar_systematic_2025,ruppel_identification_2025}. Alongside low-cost external sensors such as MEMS accelerometers~\cite{hassan_advanced_2026}, internal machine signals from the feed drives and spindle offer an alternative to dynamometers. These signals respond to the cutting process, allowing indirect force estimation, although drive dynamics also influence their response ~\cite{ruppel_identification_2025,schmucker_wide_2023}.

Feed drive signals provide directional information because each drive acts along a known machine axis. Tai and Altintas~\cite{tai_physics-guided_2026} combine current commands, velocities, and accelerations with physical relationships, while Pashmforoush et al.~\cite{pashmforoush_physics-informed_2026} combine analytically estimated forces from a feed drive model with CNC and process data. However, feed drive signals also contain contributions from inertia, friction, gravity, and other axis dynamics. Limited measurement bandwidth further restricts the resolution of rapid cutting force variations~\cite{reichel_cutting_2026}.
\begin{figure*}[!t]
  \centering
  \includegraphics[width=\textwidth]{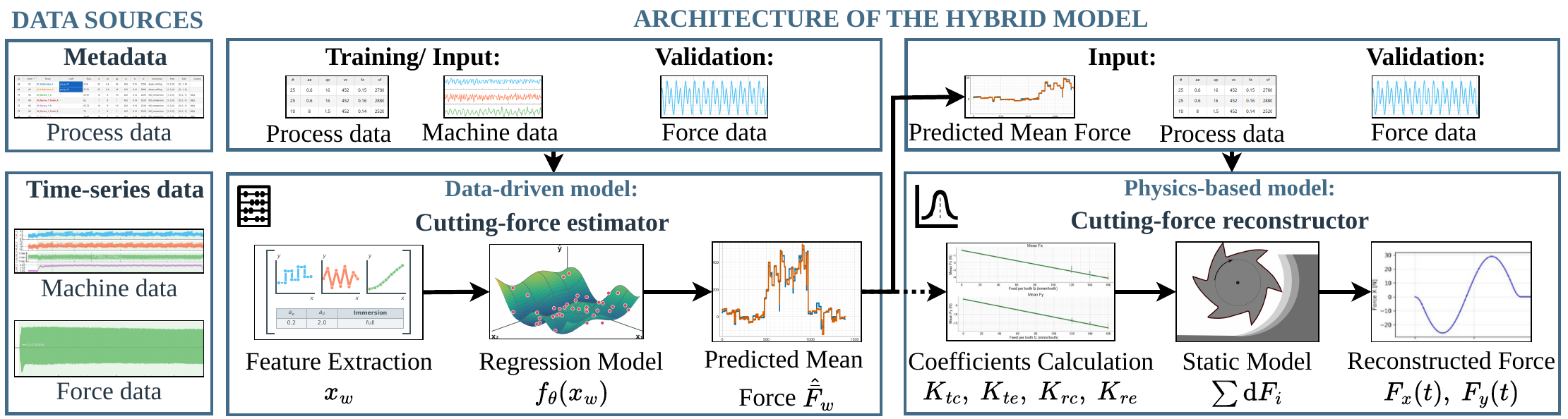}
  \caption{Overall workflow of the proposed hybrid force reconstruction approach combining internal machine signals and machining metadata with data-driven mean force estimation and physics-based reconstruction of the force progression. Dynamometer measurements are used only for development and validation.}
  \label{fig:workflow}
\end{figure*}
\noindent Spindle signals provide an alternative representation of the cutting load. During milling, the spindle drive compensates for the cutting torque to maintain the commanded rotational speed. Spindle current and related signals therefore contain information about the tangential cutting load with fewer contributions from changing axis motion. Xin et al.~\cite{xin_optimization_2026} optimized mechanistic cutting force coefficients using a genetic algorithm and reported an approximately linear relationship between spindle current and cutting load after compensation for the no-load current. Meng et al.~\cite{meng_milling_2026} predict the force progression from time-dependent spindle current signals, while relationships between spindle signals and dynamometer measurements have also been demonstrated using industrial machine data~\cite{chehrehzad_sensorless_2025}. Nevertheless, spindle-based estimation is limited by spindle dynamics, controller filtering, and the sampling rates of commonly available machine signals. Although these signals reflect changes in cutting load, they may not fully resolve high-frequency force variations.

Estimating forces from this limited information therefore requires suitable data-driven, physical, or hybrid models. For industrial deployment, low computational complexity and modest training-data requirements remain desirable. This work therefore investigates a lightweight cascade combining selected spindle and feed drive signals with machining metadata. Rather than learning the instantaneous force progression, classical regression estimates mean in-plane forces. These parameterize a static milling force model that reconstructs the angular force progression over one spindle revolution. The main contributions are:

\begingroup
\setlength{\partopsep}{0pt} %
\begin{enumerate}
    \item Lightweight data-driven estimation of mean in-plane cutting forces from spindle signals, feed drive signals and machining metadata.
    \item Hybrid reconstruction of the force progression using a data-driven approach and a static milling model.
    \item Experimental validation across varying process conditions, tools, and workpiece materials, with force sensing required only for development and validation.
\end{enumerate}
\endgroup

\section{Force estimation methodology}
\label{sec:method}
\noindent The methodology covers the hybrid workflow (Section~\ref{subsec:overall-approach}), feature extraction and selection (Section~\ref{subsec:feature-extraction}), mean force prediction (Section~\ref{subsec:force-estimator}), and force reconstruction (Section~\ref{sec:static-reconstruction}).

\subsection{Overall approach}
\label{subsec:overall-approach}
\noindent Fig.~\ref{fig:workflow} shows the overall workflow of the proposed hybrid force reconstruction. Here, machining metadata and time-series signals of the machine form the main input data, while dynamometer measurements are used only for model development and validation.

The cutting force estimator combines statistical features, signal complexity measures, and machining metadata to predict $\bar{F}_x$ and $\bar{F}_y$. It uses a regression model to map one feature vector per run to one mean force vector, keeping the learning task compact and avoiding sample-wise regression along the force curve.

The cutting force reconstructor uses these mean forces to identify cutting force coefficients. Together with tool geometry and engagement conditions, the coefficients determine the in-plane force progression over one spindle revolution.

\subsection{Data-driven model: Feature extraction}
\label{subsec:feature-extraction}

\noindent Sixteen internal controller signals were acquired using the LSV2 communication protocol. The cutting intervals were identified automatically using thresholds on the commanded spindle current. For each extracted signal segment, a broad set of statistical features and signal complexity measures was initially calculated.

For numeric features, one feature from each pair with an absolute Pearson correlation above $0.98$ was discarded. The remaining features were evaluated using an Extra-Trees regressor and permutation importance with negative mean absolute error as the scoring metric. Features from machine signals and machining metadata were ranked separately. Within each group, features with positive importance were retained until 95\% of the cumulative importance was reached.
The same dataset was used for feature selection and cross-validation.

This feature selection procedure kept 11 features from four spindle signals and one feed drive signal (Table~\ref{tab:selected-signal-features}).
\begin{table}[t]
  \caption{Selected machine signals and their features used as model inputs.}
  \label{tab:selected-signal-features}
  \centering
  \small
  \setlength{\tabcolsep}{3pt}
  \renewcommand{\arraystretch}{0.9}
  \begin{tabular}{@{}lccc@{}}
  \toprule
  Signal & Mean & HFD$_{\mathrm{med}}$ & SVD$_{\mathrm{med}}$ \\
  \midrule
  Electrical spindle power & x & x & x \\
  Actual spindle torque & x & x & x \\
  Actual spindle speed & x & x & x \\
  Commanded spindle current & x & -- & -- \\
  Actual feed-axis velocity & x & -- & -- \\
  \bottomrule
\end{tabular}

\end{table}
The selected features include the global mean, median Higuchi fractal dimension (HFD)~\cite{higuchi_approach_1988}, and median normalized singular-value-decomposition (SVD) entropy~\cite{sandoval_entropy_2021}.

For each signal segment $s$, HFD and normalized SVD entropy were evaluated in $W$ non-overlapping windows $s_w$ of $0.333\,\mathrm{s}$. Each segment was represented by the median value across its windows,
\begin{equation}
  \widetilde{q}_s=
  \operatorname{median}_{w=1,\ldots,W} q(s_w),
  \qquad
  q\in\{\mathrm{HFD},H_{\mathrm{SVD}}\}.
  \label{eq:windowed-features}
\end{equation}
The same selection procedure retained 14 process and setup metadata descriptors (Table~\ref{tab:selected-metadata-features}).
\begin{table}[t]
  \caption{Selected process and setup metadata used as model inputs.}
  \label{tab:selected-metadata-features}
  \centering
  \small
  \setlength{\tabcolsep}{3pt}
  \renewcommand{\arraystretch}{0.9}
  \begin{tabular}{@{}ll@{}}
  \toprule
  Group & Selected metadata \\
  \midrule
  Cutting & Radial engagement $a_e$, axial depth of cut $a_p$ \\
          & Cutting speed $v_c$, feed per tooth $f_z$ \\
  Tool & Identity, diameter, material, coating \\
  Workpiece & Material \\
  Process & Cutting modes, cooling mode \\
  Motion & Feed-direction sign, shift-direction components ($y$, $z$) \\
  \bottomrule
\end{tabular}

\end{table}
\subsection{Data-driven model: Cutting force estimator}
\label{subsec:force-estimator}

\noindent The cutting force estimator predicts run-level mean forces to parameterize the static model. It maps the combined features from machine signals and machining metadata $\boldsymbol{x}_w$ to the mean in-plane components:
\begin{equation}
  \hat{\bar{\boldsymbol{F}}}_w =
  f_{\boldsymbol{\theta}}(\boldsymbol{x}_w), \qquad
  \hat{\bar{\boldsymbol{F}}}_w =
  [\hat{\bar{F}}_x,\hat{\bar{F}}_y]^\mathsf{T}.
  \label{eq:ml}
\end{equation}
Fourteen classical regressors were evaluated using identical grouped cross-validation partitions. Ridge ($\alpha=1$) achieved the lowest out-of-fold MAE for both in-plane components and was selected as the primary model. After preprocessing, Ridge evaluates linear combinations of the $n$ encoded inputs, requiring $\mathrm{O}(n)$ operations per run for the two force outputs~\cite{sierotowicz_emg-driven_2022}. Four additional models were retained to assess how the regression algorithm influences the subsequent force reconstruction. Hyperparameters were specified in advance rather than automatically optimized to avoid additional data-dependent tuning and reduce the risk of overfitting in this initial study.

Numerical features were standardized, while categorical features were one-hot encoded.  Scaling and encoding were fitted exclusively on the respective training fold. Five-fold grouped cross-validation kept runs from the same physical setup together.

\subsection{Physics-based model: Cutting force reconstruction}
\label{sec:static-reconstruction}

\noindent The predicted mean forces are used to identify cutting force coefficients and reconstruct the in-plane force progression. Both steps are based on the milling force formulation of Altintas~\cite{altintas_manufacturing_2012}. The model uses four in-plane cutting coefficients: $K_{tc}$ and $K_{rc}$ describe the chip-thickness-dependent tangential and radial force contributions, while $K_{te}$ and $K_{re}$ describe the corresponding chip-thickness-independent edge contributions.

For a fixed cutter geometry and axial depth of cut, the in-plane cutting coefficients are identified from the predicted mean forces. For full immersion,
\begin{align}
  \bar{F}_x &= -\frac{Z a_p}{4}K_{rc}f_z-\frac{Z a_p}{\pi}K_{re},\notag\\
  \bar{F}_y &= \phantom{-}\frac{Z a_p}{4}K_{tc}f_z+\frac{Z a_p}{\pi}K_{te},
  \label{eq:full-immersion-means}
\end{align}
where $Z$ is the number of flutes, $a_p$ the axial depth of cut, and $f_z$ the feed per tooth. For fixed $Z$ and $a_p$, Eq.~\ref{eq:full-immersion-means} is linear in $f_z$, with slopes $m_x,m_y$ and offsets $b_x,b_y$:
\begin{equation}
  \bar{F}_x=m_xf_z+b_x, \qquad
  \bar{F}_y=m_yf_z+b_y
  \label{eq:mean-force-fits}
\end{equation}
The cutting and edge coefficients are then calculated from these slopes and offsets as
\begin{align}
  K_{rc} &= -\frac{4m_x}{Z a_p}, &
  K_{re} &= -\frac{\pi b_x}{Z a_p},\notag\\
  K_{tc} &= \phantom{-}\frac{4m_y}{Z a_p}, &
  K_{te} &= \phantom{-}\frac{\pi b_y}{Z a_p}.
  \label{eq:coefficient-identification}
\end{align}
Identification requires at least two distinct feeds per tooth with the remaining cutting conditions fixed. For side milling, the coefficients are obtained by least-squares fitting of the analytical mean force equations, accounting for the entry and exit angles.
Errors in the predicted mean forces can yield physically implausible coefficients. A physics-guided correction therefore reverses force--feed slope signs that imply negative cutting coefficients. Full immersion additionally uses the study-specific, empirically chosen constraint $K_{tc}/K_{rc}\geq3$ and low-force slope limits to avoid implausible edge contributions. Updated offsets $b_i^*=\operatorname{mean}(\bar F_i)-m_i^*\operatorname{mean}(f_z)$, $i\in\{x,y\}$, preserve the mean force level. Here, $\operatorname{mean}$ averages over fitting points and asterisks denote corrected values.

The static milling force model then uses the corrected coefficients to reconstruct the forces. It determines the engaged parts of each helical tooth, calculates their local forces, transforms them into machine coordinates, and sums their contributions.
The cutter of radius $R$ and helix angle $\beta$ is divided into axial slices of height $\mathrm{d}z$. For tooth $j$ and slice $k$ at axial position $z_k$, the angular position at time $t$ is
\begin{equation}
  \varphi_{j,k}(t)=
  \omega t+\frac{2\pi(j-1)}{Z}
  -\frac{z_k\tan\beta}{R},
  \label{eq:helical-angle}
\end{equation}
where $\omega$ is the spindle angular velocity. Since engagement repeats every revolution, tooth angles are mapped to $[0,2\pi)$ using $\tilde{\varphi}_{j,k}=\varphi_{j,k}\bmod 2\pi$. A slice generates force only between the entry angle $\varphi_{\mathrm{en}}$ and exit angle $\varphi_{\mathrm{ex}}$:
\begin{equation}
  \varphi_{\mathrm{en}}\leq\tilde{\varphi}_{j,k}\leq\varphi_{\mathrm{ex}},\qquad \varphi_{\mathrm{en}},\varphi_{\mathrm{ex}}\in[0,\pi].
  \label{eq:engagement-condition}
\end{equation}
The radial engagement $a_e$ determines the engagement angle $\theta_e=\arccos(1-a_e/R)$. In the adopted angular convention, cutting occurs from $\pi-\theta_e$ to $\pi$ for down milling and from $0$ to $\pi$ for full immersion.

For each engaged slice, the undeformed chip thickness $h_{j,k}$ and the corresponding tangential and radial force contributions are
\begin{align}
  h_{j,k} &= f_z\sin(\tilde{\varphi}_{j,k}),\notag\\
  \mathrm{d}F_{t,j,k}
  &= [K_{tc}h_{j,k}+K_{te}]\,\mathrm{d}z,\notag\\
  \mathrm{d}F_{r,j,k}
  &= [K_{rc}h_{j,k}+K_{re}]\,\mathrm{d}z.
  \label{eq:staticmodel}
\end{align}
These local force contributions are transformed into the machine $x$-$y$ coordinate system:
\begin{align}
  \mathrm{d}F_{x,j,k}
  &= -\mathrm{d}F_{t,j,k}\cos(\tilde{\varphi}_{j,k})
     -\mathrm{d}F_{r,j,k}\sin(\tilde{\varphi}_{j,k}),\notag\\
  \mathrm{d}F_{y,j,k}
  &= \phantom{-}\mathrm{d}F_{t,j,k}\sin(\tilde{\varphi}_{j,k})
     -\mathrm{d}F_{r,j,k}\cos(\tilde{\varphi}_{j,k}).
  \label{eq:in-plane-transformation}
\end{align}
Finally, the contributions of all engaged teeth and axial slices are summed to obtain the instantaneous in-plane forces:
\begin{equation}
  F_x(t)=\sum_{(j,k)\,\mathrm{engaged}}\mathrm{d}F_{x,j,k},
  \qquad
  F_y(t)=\sum_{(j,k)\,\mathrm{engaged}}\mathrm{d}F_{y,j,k}.
  \label{eq:force-summation}
\end{equation}

\section{Experimental setup}
\label{sec:setup}

\noindent The experimental setup covers the machine, cutting conditions, and data acquisition (Section~\ref{subsec:data-acquisition}), followed by the evaluation with setup grouping and cross-validation (Section~\ref{subsec:evaluation-protocol}).

\subsection{Machine tool, cutting process, and data acquisition}
\label{subsec:data-acquisition}

\noindent Experiments were conducted on a five-axis Hermle C40 machining center with a Heidenhain iTNC 530 control. Two workpiece materials were investigated: quenched-and-tempered 42CrMo4+QT steel and AlMg4.5Mn aluminium alloy. Eight solid-carbide end mills were used, covering diameters of $6$--$12\,\mathrm{mm}$, one to four cutting edges, helix angles of $30^\circ$--$50^\circ$, and uncoated and TiAlN-coated variants. The investigated cutting conditions are summarized in Table~\ref{tab:conditions}.

Controller signals were acquired on an external computer using the LSV2 communication protocol over a WLAN TCP/IP connection. The controller interface allowed simultaneous acquisition of 16 signals. Four signals from the Controller Computer (CC) were sampled at $1666.7\,\mathrm{Hz}$: electrical spindle power, actual spindle torque, actual spindle speed, and commanded spindle current. The remaining 12 signals originated from the interpolator (IPO) and were sampled at $333.3\,\mathrm{Hz}$. These included the path feed rate, actual and commanded positions of the $x$-, $y$-, and $z$-axes, actual velocities of all three axes, and commanded velocities of the $x$- and $y$-axes.

\begin{figure}[!tbp]
  \centering
  \includegraphics[width=\linewidth]{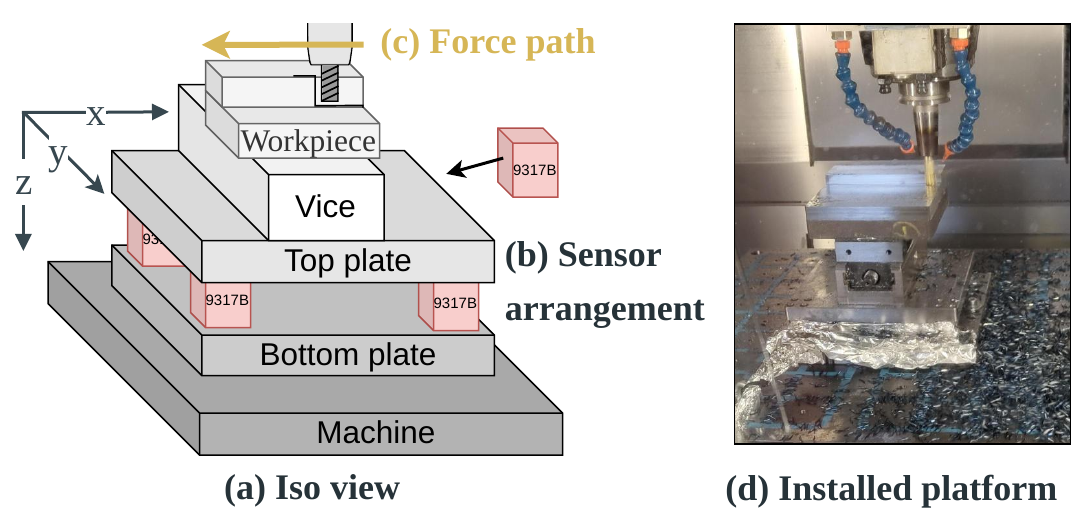}
  \caption{Reference-force measurement setup: (a) isometric view, (b) arrangement of four Kistler Type 9317B force links, (c) cutting force load path, and (d) installed platform.}
  \label{fig:setup}
\end{figure}

Reference forces were measured with a custom three-component platform (Fig.~\ref{fig:setup}). The vice holding the workpiece was mounted on the platform's top plate. Four Kistler Type 9317B piezoelectric force sensors were installed between this plate and the machine-mounted base plate. A Kistler Type 5431 summing box combined the sensor outputs. A Type 5070 multichannel charge amplifier conditioned them for acquisition at $30\,\mathrm{kHz}$ by an NI 9215 module in a cDAQ-9178 chassis.
Without a common hardware clock, computer timestamps provided initial synchronization with approximately $100$--$400\,\mathrm{ms}$ uncertainty. Post-processing refined alignment at tool entry using the corresponding rises in measured force and spindle power.

\begin{table}[t]
  \caption{Cutting conditions and dataset composition.}
  \label{tab:conditions}
  \centering
  \small
  \renewcommand{\arraystretch}{0.9}
  \begin{tabular}{@{}lcl@{}}
  \toprule
  Quantity & Symbol & Values or range \\
  \midrule
  Spindle speed & $n$ & $3{,}981$--$18{,}000\,\mathrm{min}^{-1}$ \\
  Feed per tooth & $f_z$ & $0.013$--$0.192\,\mathrm{mm}$ \\
  Axial depth of cut & $a_p$ & $0.3$--$17\,\mathrm{mm}$ \\
  Radial engagement & $a_e$ & $0.16$--$8\,\mathrm{mm}$ \\
  Cutting modes & -- & Down milling; full immersion \\
  Tool condition & -- & Fresh/unworn \\
  Runs / setup groups & -- & $1{,}391$ / $74$ \\
  Grouped CV folds & -- & $5$ \\
  \bottomrule
\end{tabular}

\end{table}
\subsection{Evaluation protocol}
\label{subsec:evaluation-protocol}

\noindent The dataset comprised 1,391 labeled cutting runs acquired with fresh, unworn tools. Tool wear was not varied experimentally. Because the dataset included repeated cutting configurations, random run-level splitting could produce overly optimistic performance estimations.
The runs were therefore organized into 74 setup groups. Each group contains runs with the same complete physical configuration, including process conditions, tool and workpiece properties, cutting mode, motion conditions, cooling, and trend configuration. Any change in these properties defines a new setup group. Five-fold grouped cross-validation kept each setup group entirely within training or testing. Each group was used for testing exactly once, and all 14 regressors used identical folds.

Performance was evaluated from the combined out of fold predictions using mean absolute error (MAE), root mean square error (RMSE), and the coefficient of determination $R^2$. Based on these results, Bayesian Ridge, automatic relevance determination (ARD) regression, Ridge, Elastic Net, and Extra Trees were retained for further comparison.

Reconstruction was evaluated on 847 runs with known tool helix geometry. Reconstructed and measured forces were expressed in the same coordinate system and force-sign convention for comparison. To reduce noise, measurements were low-pass filtered using a zero-phase Butterworth filter with a tooth-passing-frequency-dependent cutoff and angularly aligned with the reconstructions. The $R^2$ value was calculated from the corresponding force values across all evaluated tool revolutions.
\section{Results and discussion}
\label{sec:results}

\noindent The results evaluate mean force prediction (Section~\ref{subsec:mean-force-results}) and static force reconstruction (Section~\ref{subsec:reconstructed-force-results}).
\subsection{Prediction of mean forces}
\label{subsec:mean-force-results}
\begin{figure}[!tbp]
  \centering
  \includegraphics[width=\linewidth,trim=0 3.2bp 0 3.9bp,clip]{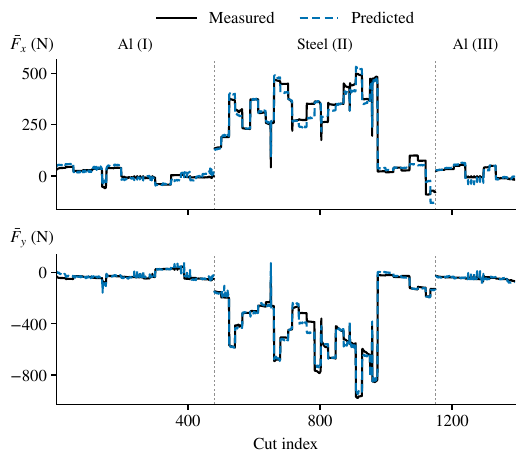}
  \caption{Measured and predicted mean forces. Segments I and III use AlMg4.5Mn aluminium, and segment II uses 42CrMo4+QT steel.}
  \label{fig:mean-results}
\end{figure}

\noindent The five best models for mean force estimation are compared in Table~\ref{tab:results}.
\begin{table}[!htb]
  \caption{Grouped out-of-fold mean force performance of the five shortlisted models. MAE and RMSE are given in N. Best results are bold.}
  \label{tab:results}
  \centering
  \scriptsize
  \setlength{\tabcolsep}{7.5pt}
  \renewcommand{\arraystretch}{0.9}
  \begin{tabular}{@{}lcccccc@{}}
  \toprule
  & \multicolumn{3}{c}{$\bar{F}_x$} & \multicolumn{3}{c}{$\bar{F}_y$} \\
  \cmidrule(lr){2-4}\cmidrule(l){5-7}
  Model & MAE & RMSE & $R^2$ & MAE & RMSE & $R^2$ \\
  \midrule
  Ridge & $\mathbf{15.0}$ & $\mathbf{21.1}$ & $\mathbf{0.984}$ & $\mathbf{15.0}$ & $\mathbf{23.0}$ & $\mathbf{0.991}$ \\
  ARD regression & $17.0$ & $24.9$ & $0.978$ & $16.0$ & $32.0$ & $0.982$ \\
  Bayesian Ridge & $17.2$ & $25.0$ & $0.978$ & $15.9$ & $31.3$ & $0.983$ \\
  Elastic Net & $20.0$ & $28.2$ & $0.972$ & $24.3$ & $33.9$ & $0.980$ \\
  Extra Trees & $18.4$ & $29.9$ & $0.968$ & $26.2$ & $47.1$ & $0.962$ \\
  \bottomrule
\end{tabular}

\end{table}
Ridge achieved the lowest MAE of $15.0\,\mathrm{N}$ for both $\bar{F}_x$ and $\bar{F}_y$, with corresponding $R^2$ values of $0.984$ and $0.991$. Figure~\ref{fig:mean-results} compares the predicted and measured mean forces across the test folds.

The predictions follow the measured force trends over the investigated force range, although systematic offsets occur for some cutting configurations. The model also captures the substantially different force levels observed for steel and aluminium. Overall, the results indicate that the mean in plane forces can be estimated reliably from the selected machine signals and machining metadata.

\subsection{Reconstructed milling forces}
\label{subsec:reconstructed-force-results}

\noindent The static model is evaluated using mean forces predicted by regression or calculated from force sensor measurements. Model names refer to prediction-based reconstructions, while the baseline is the sensor-based reconstruction. Both are compared with each other and with filtered force measurements (Table~\ref{tab:static-results}).
\begin{figure*}[!tp]
  \centering
  \begin{minipage}[c]{0.48\textwidth}
    \centering
    \includegraphics[width=\linewidth,trim=0 0 0 80bp,clip]{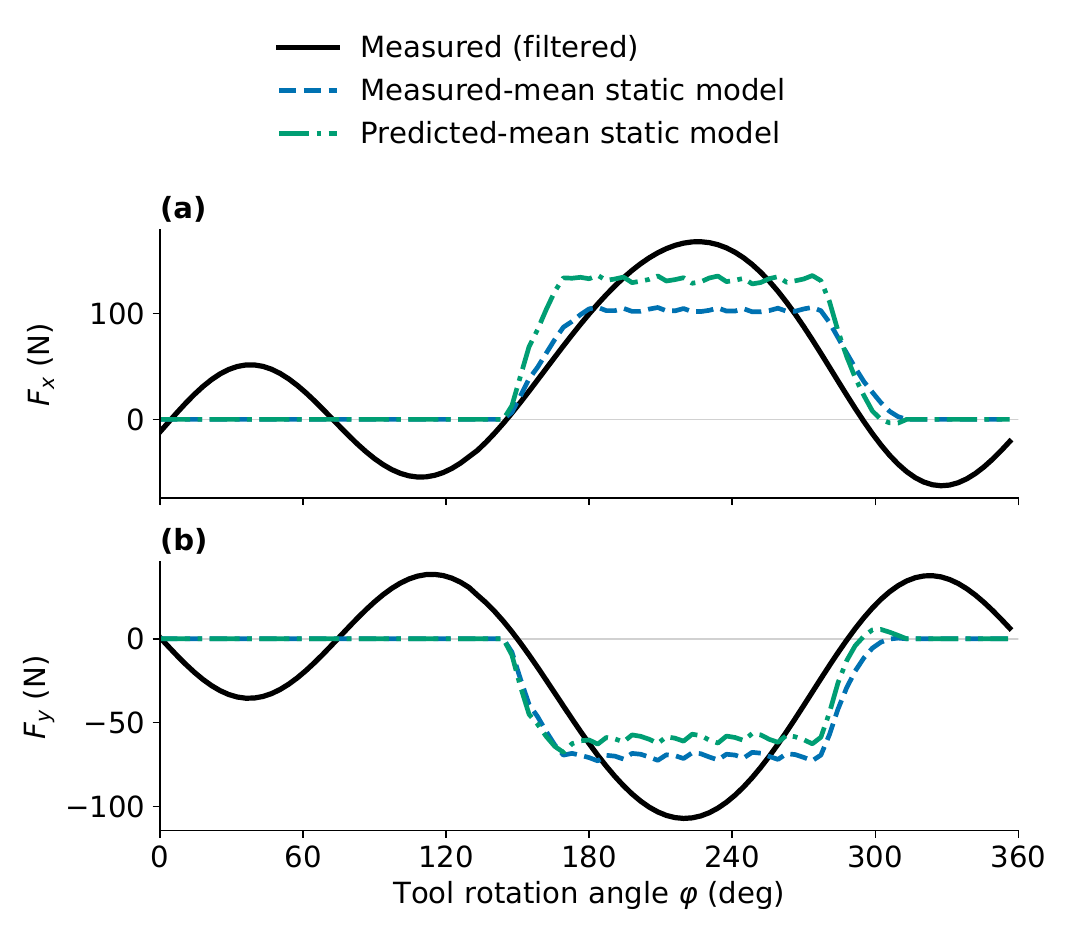}
    \par\smallskip{\footnotesize (I)}
  \end{minipage}\hfill
  \begin{minipage}[c]{0.48\textwidth}
    \centering
    \includegraphics[width=\linewidth,trim=0 0 0 80bp,clip]{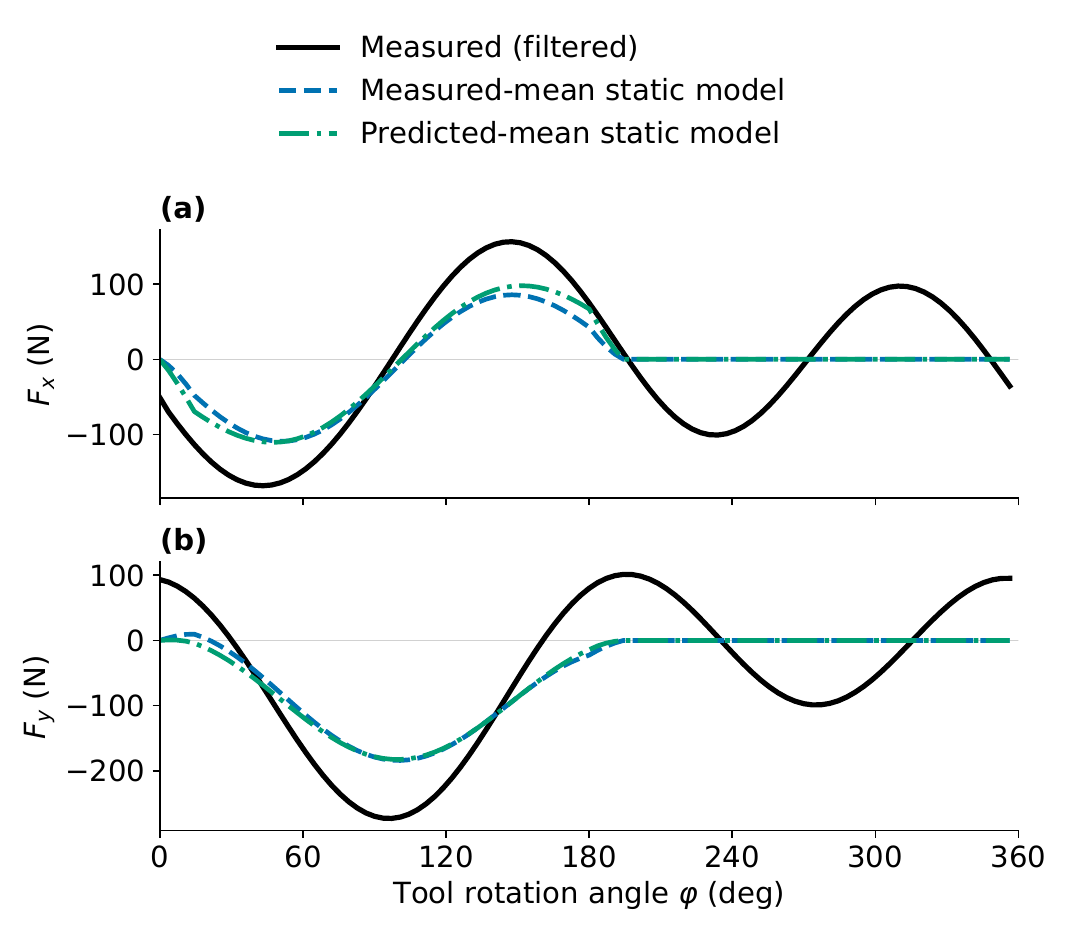}
    \par\smallskip{\footnotesize (II)}
  \end{minipage}

  \medskip

  \makebox[\textwidth][c]{%
    \begin{minipage}[c]{0.48\textwidth}
      \centering
      \includegraphics[width=\linewidth,trim=0 0 0 80bp,clip]{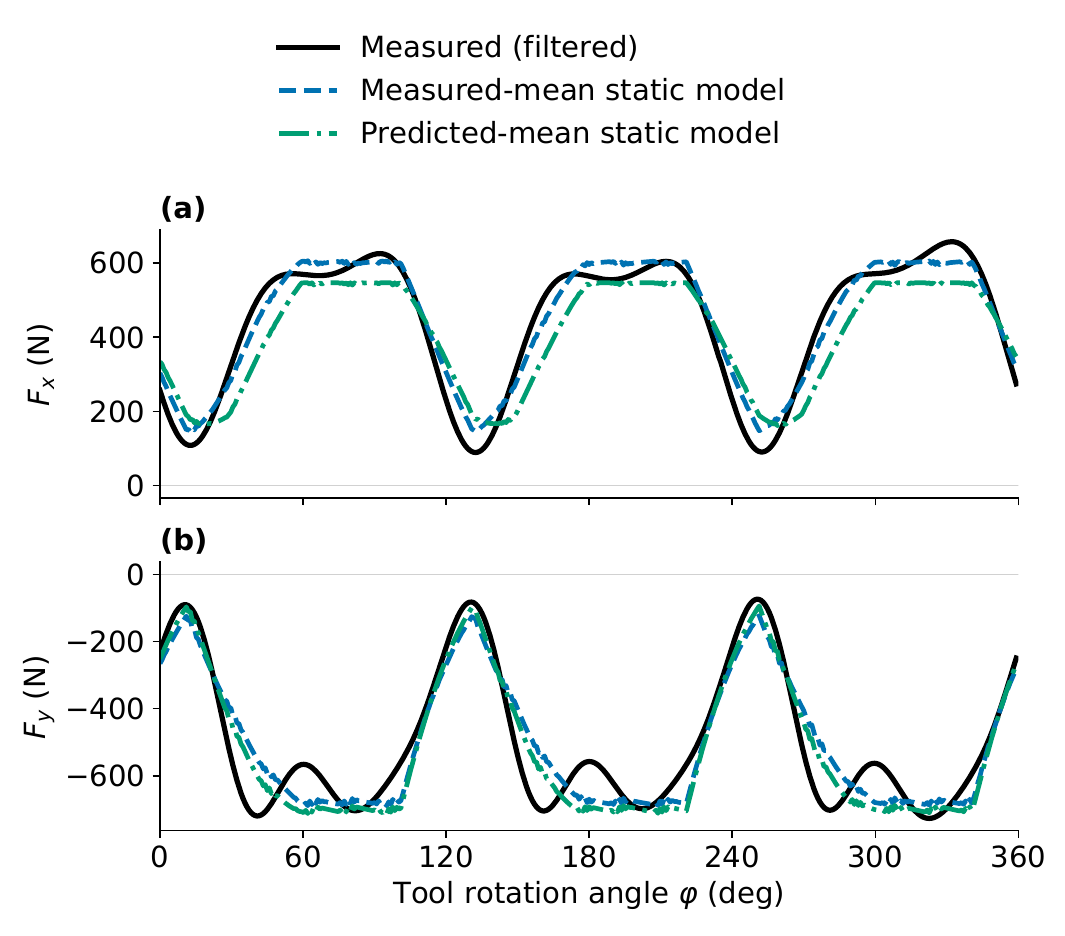}
      \par\smallskip{\footnotesize (III)}
    \end{minipage}%
    \hspace{0.025\textwidth}%
    \begin{minipage}[c]{0.192\textwidth}
      \centering
      \includegraphics[width=\linewidth]{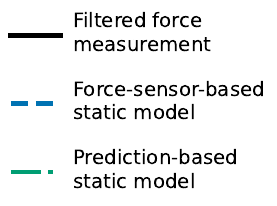}
    \end{minipage}%
  }
  \caption{Representative $F_x$ (a) and $F_y$ (b) profiles: (I) aluminium side milling, $6\,\mathrm{mm}$ single-flute cutter. (II) aluminium full-slot milling, $8\,\mathrm{mm}$ single-flute cutter. (III) steel milling, three-flute cutter. Filtered measurements and static reconstructions use the shared legend.}
  \label{fig:profiles}
\end{figure*}
\begin{table}[!htb]
  \caption{Comparison of prediction-based and sensor-based static models with each other and with filtered force measurements using $R^2$. Best values are bold.}
  \label{tab:static-results}
  \centering
  \scriptsize
  \setlength{\tabcolsep}{7pt}
  \renewcommand{\arraystretch}{0.9}
  \begin{tabular}{@{}lcccc@{}}
  \toprule
  & \multicolumn{2}{c}{$F_x$} & \multicolumn{2}{c}{$F_y$} \\
  \cmidrule(lr){2-3}\cmidrule(l){4-5}
  \shortstack[l]{Static model\\based on} & \shortstack{Pred. static/\\Sensor static} & \shortstack{Static/\\force signal} & \shortstack{Pred. static/\\Sensor static} & \shortstack{Static/\\force signal} \\
  \midrule
  Ridge & $\mathbf{0.962}$ & 0.506 & 0.982 & 0.625 \\
  ARD regression & 0.933 & 0.501 & 0.949 & 0.604 \\
  Bayesian Ridge & 0.923 & 0.500 & 0.951 & 0.606 \\
  Elastic Net & 0.904 & 0.492 & $\mathbf{0.986}$ & $\mathbf{0.629}$ \\
  Extra Trees & 0.925 & 0.449 & 0.980 & 0.626 \\
  \midrule
  Sensor baseline & -- & $\mathbf{0.531}$ & -- & 0.627 \\
  \bottomrule
\end{tabular}

\end{table}
Compared with the sensor-based baseline, Ridge reaches $R^2=0.962$ for $F_x$ and Elastic Net $0.986$ for $F_y$. Against filtered force measurements, the baseline reaches $0.531$ and $0.627$ for $F_x$ and $F_y$, respectively, versus Ridge's $0.506$ and $0.625$. Using predicted rather than sensor-derived mean forces adds relatively little discrepancy in this pooled evaluation. 

Model rankings change for $F_y$ between predicted means and force reconstruction. Ridge predicts mean forces more accurately, but Elastic Net slightly outperforms Ridge for $F_y$ reconstruction with $R^2=0.986$ versus $0.982$. This suggests that mean force errors affect the force feed relationships and the resulting coefficients differently and therefore do not translate directly into reconstruction errors. In addition, the physics based correction modifies this error propagation by enforcing physically plausible coefficient constraints, although it cannot eliminate it completely.

Figure~\ref{fig:profiles} complements the numerical comparison with representative force curves. In aluminium side milling (Fig.~\ref{fig:profiles}, I), the static model does not reproduce the detailed curve shape during engagement, but captures the force level and engagement interval. The two static models produce comparable results. In aluminium full-slot milling (Fig.~\ref{fig:profiles}, II), both reconstructions reproduce the curve shape and engagement behavior well, although the force level is slightly underestimated. The steel example (Fig.~\ref{fig:profiles}, III) shows the closest correspondence in force level and curve shape. This may partly result from the higher cutting forces, for which noise and unmodeled disturbances have a smaller relative influence.

Across all three conditions, the prediction-based static model closely follows the force-sensor-based model, supporting reconstruction from machine signals without force measurements during deployment. Of course, the static model cannot reproduce local peaks, oscillations, and other dynamic components of the measured signal. Nevertheless, it captures the principal force direction, load level, dominant engagement-related peaks, and periodic engagement patterns relevant to process monitoring.

\section{Conclusion and outlook}
\label{sec:conclusion}

\noindent A regression model estimates one mean force vector from internal machine signals and machining metadata, keeping the learning task compact and avoiding regression at every point of the force curve. A static milling force model then uses this force vector to reconstruct quasi-static force curves, eliminating the need for dynamometer measurements during deployment.

For mean force estimations, Ridge achieved $R^2=0.984$ and $0.991$ for $\bar{F}_x$ and $\bar{F}_y$, with an MAE of $15.0\,\mathrm{N}$ for both. Its static reconstructions reached $R^2=0.962$ and $0.982$ against the force-sensor model. Against force measurements, Ridge reached $R^2=0.506$ and $0.625$, compared with $0.531$ and $0.627$ for the force-sensor baseline. These results show that mean forces predicted from machine signals can replace sensor-derived mean forces as inputs to the static model, with little loss of reconstruction accuracy. Model rankings changed after reconstruction, showing that mean force accuracy alone is insufficient for selecting the complete cascade. The static model captures principal load levels, engagement intervals, and angular force progression without explicit dynamic modeling, providing a basis for future process monitoring and optimization. However, it does not reproduce dynamic oscillations or local peaks, limiting its suitability for milling applications where these effects are essential.

This study covered multiple unworn tools on one machine. Further validation on other machines, with additional tool geometries and under progressive wear, would strengthen the evidence for transferability. Online reconstruction for process monitoring and optimization remains future work.
\section*{Funding}
\noindent This work was funded by the State of Baden-Württemberg through the Ministry of Science, Research and the Arts under the program “Innovative Projekte / Kooperationsprojekte” for the project PATTERN.
\section*{Data Availability} 
\noindent Data generated during the current study are available from the corresponding author on reasonable request.

\section*{Declaration of generative AI and AI-assisted technologies in the writing process}
\noindent The authors used ChatGPT, OpenAI Codex, and Google Gemini for language refinement and manuscript organization. They reviewed and edited the output and take full responsibility for the article.

\bibliographystyle{IEEEtran}
\bibliography{references}

\end{document}